\documentclass[11pt,a4paper]{article}
\usepackage{authblk}
\usepackage[hyperref]{acl2021}
\usepackage{times}
\usepackage{latexsym}

\usepackage{microtype}
\usepackage{amsthm}
\usepackage{amssymb}
\usepackage{graphicx}
\newtheorem{definition}{Definition}
\usepackage[english]{babel}
\usepackage[autostyle, english = american]{csquotes}
\MakeOuterQuote{"}
\newcommand*\mean[1]{\overline{#1}}
\usepackage{multirow,makecell}
\usepackage{diagbox}
\usepackage{slashbox}
\usepackage{pgfplots}
\pgfplotsset{compat=1.17}
\usepackage{caption}
\usepackage{subcaption}
\usepackage{url}
\usepackage{booktabs}
\usepackage{natbib}
\usepackage{siunitx}

\aclfinalcopy % Uncomment this line for the final submission
\makeatletter
\newcommand{\printfnsymbol}[1]{%
  \textsuperscript{\@fnsymbol{#1}}%
}
\makeatother

\title{Switching Contexts: Transportability Measures for NLP}

\author{Guy Marshall\thanks{\ \ \ \texttt{equal contribution}}$~^{\dagger}$, Mokanarangan Thayaparan\printfnsymbol{1}$^{\dagger}$, Philip Osborne$^{\dagger}$, Andr\'e Freitas$^{\dagger}$$^{\ddagger}$ \\  Department of Computer Science, University of Manchester, United Kingdom$^{\dagger}$ \\  Idiap Research Institute, Switzerland$^{\ddagger}$ \\ \texttt{\{guy.marshall, philip.osborne\}@postgrad.manchester.ac.uk} \protect\\ \texttt{\{mokanarangan.thayaparan, andre.freitas\}@manchester.ac.uk} }

\date{}

\begin{document}
\maketitle

\begin{abstract}
    This paper explores the topic of transportability, as a sub-area of generalisability. By proposing the utilisation of metrics based on well-established statistics, we are able to estimate the change in performance of NLP models in new contexts. Defining a new measure for transportability may allow for better estimation of NLP system performance in new domains, and is crucial when assessing the performance of NLP systems in new tasks and domains. Through several instances of increasing complexity, we demonstrate how lightweight domain similarity measures can be used as estimators for the transportability in NLP applications. The proposed transportability measures are evaluated in the context of Named Entity Recognition and Natural Language Inference tasks.
\end{abstract}

\section{Introduction}

The empirical evaluation of the quality of NLP models under a specific task is a fundamental part of the scientific method of the NLP community. However, commonly, many proposed models are found to perform well in the specific context in which they are evaluated and state-of-the-art claims are usually found not transportable to similar but different settings. The current evaluation metrics may only indicate which algorithm or setup performs best: they are unable to estimate performance in a new context, to demonstrate internal validity, or to verify causality. To offset this, statistical significance testing is sometimes applied in conjunction with performance measures (e.g. F1-score, BLEU) to attempt to establish validity. However, statistical significance testing has been shown to be lacking. \citet{dror2018hitchhiker} reviewed NLP papers from ACL17 and TACL17 and found that only a third of these papers use significance testing. Further, many papers did not specify the type of test used, and some even employed an inappropriate statistical test.

Performance is measured in NLP tasks primarily through F1 score or task-specific metrics such as BLEU. The limited scope of these as performance evaluation techniques has been shown to have issues. \citet{sogaard2014s} highlights the data selection bias in NLP system performance. \citet{gorman2019we} show issues of using standard splits, as opposed to random splits. We support their statement that "practitioners who wish to firmly establish that a new system is truly state-of-the-art augment their evaluations with Bonferroni-corrected random split hypothesis testing". In an NLI task, using SNLI and MultiNLI datasets with a set of different models, it has been shown that permutations of training data leads to substantial changes in performance \citep{schluter2018data}.

Further, the lack of transportability for NLP tasks has been raised by specialists in applied domains. For example, healthcare experts have expressed their frustration in the limitations of algorithms built in research settings for practical applications \citep{demner2016aspiring} and the reduction of performance ``outside of their development frame'' \citep{maddox2015natural}. More generally, ``machine learning researchers have noted current systems lack the ability to recognize or react to new circumstances they have not been specifically programmed or trained for'' \citep{pearl2019seven}. 

The advantages of "more transportable" approaches, such as BERT, in terms of their performance in multiple different domains, is currently not expressed (other than the prevalence of such architectures across a range of state-of-the-art tasks and domains).
To support analysis and investigation into the insight that could be gained by examination of these properties, we suggest metrics and a method for measuring the transportability of models to new domains. This has immediate relevance for domain experts, wishing to implement existing solutions on novel datasets, as well as for NLP researchers wishing to assemble new dataset, design new models, or evaluate approaches.

To support this, we propose feature gradient, and show it to have promise as a way to gain lexical or semantic insight into factors influencing the performance of different architectures in new domains. This differs from data complexity, being a comparative measure between two datasets. We aim to start a conversation about evaluation of systems in a broader setting, and to encourage the creation and utilisation of new datasets.

This paper focuses on the design and evaluation of a lightweight transportability measure in the context of the empirical evaluation of NLP models. A further aim is to provide a category of measures which can be used to estimate the stability of the performance of a system across different domains. An initial transportability measure is built by formalising properties of performance stability and variation under a statistical framework. The proposed model is evaluated in the context of Named Entity Recognition tasks (NER) and Natural Language Inference (NLI) tasks across different domains.

Our contribution is to present a measure that evaluates the transportability and robustness of an NLP model, to evaluate domain similarity measures to understand and anticipate the transportability of an NLP model, and to compare state of the art models across different datasets for NER and NLI.

%When using an approach trained on one dataset, it is often not clear how well that same trained model would perform on a new unseen dataset. The ability to have even a good guess as to whether a model is likely to perform well on a new dataset would be useful for practitioners and domain researchers alike, and even potentially for AI researchers in identifying approaches to further explore. 
%Concretely, if we have an established NER model, trained and performing state-of-the-art on CoNLL-2003, how well would we expect it to perform on the wikipedia corpus, or WNUT? Different models, extracting different features, will differ in their performance: Can we use the difference in the data and how the models perform to gain insight into the way the model is working and the features it is identifying?

%Imagine a situation where a new dataset (say WNUT, containing twitter data) needs to be NER tagged. There are existing pre-trained CoNLL-2003 models available, such as Stanford, SpacCy, and Allen NLP. We might also have information of their performance on other datasets, such as Wiki. Given some properties of the data, can we guess which will perform better *I think we need significance test or more data for this to be solid narrative, because essentially we have only a couple of datapoints for each claim and little other evidence. 

\section{Relevant background and related work}

\subsection{Terminology}
To quote \citet{campbell2015experimental}, "External validity asks the question of generalizability: To what populations, settings, treatment variables, and measurement variables can this effect be generalized?". For \citet{pearl2014external}, transportability is how generalisable an experimentally identified causal effect is to a new population where only observational studies can be conducted. "However, there is an important difference, not often distinguished, between what might be called the potential (or generic) transferability of a study and its actual (or specific) transferability to another policy or practice decision context at another time and place." \citep{walker2010generalisability} % (Walker et al, 2010 p57) 

\citet{bareinboim2013general} explore transfer of causal information, culminating in an algorithm for identifying transportable relations. Transportability in this sense does not permit retraining in the new population, and guides our choices in this paper. Other definitions of transfer learning allow for training of the model in the new context \citep{pan2010survey}, or highlight the distinction between evidential knowledge and causal assumptions \citep{singleton2014transfer}.

\subsection{Transportability: Models evaluated across different datasets}
\citet{rezaeinia2019sentiment} consider improving transportability by demonstrating word embeddings' accuracy degrades over different datasets, and propose an algorithmic method for improved word embeddings by using word2vec, adding gloVe when missing, and filling any further missing values with random entries.
In a medical tagging task, \citet{ferrandez2012generalizability} used different train/test datasets, and compared precision and recall with self-trained vs transported-trained, finding that some tag-categories performed better than others. They postulate that degradation differences were due to the differing prevalence of entities in the transported training data. Another term from this domain is "portability", in the sense that a model could be successfully used with consideration of implementation issues such as different data formats and target NLP vocabularies \citep{carroll2012portability}.
\citet{blitzer2007biographies} created a multi-domain dataset for sentiment analysis, and propose a measure of domain similarity for sentiment analysis based on the distance between the probability distributions in terms of characteristic functions of linear classifiers. %Let A be the family of subsets of Rk corresponding to characteristic functions of linear classifiers (sets on which a linear classifier returns positive value). Then the A distance between two probability distributions is dA(D,D0) = 2 supA∈A |PrD [A] − PrD0 [A]| .

In image processing, domain transfer is an active area of research. \citet{pan2010domain} propose transfer component analysis as a method to learn subspaces which have similar data properties and data distributions in different domains. They state that domain adaptation is "a special setting of transfer learning which aims at transferring shared knowledge across different but related tasks or domains". In computer vision, \citet{peng2019moment} combine multiple datasets into a larger dataset DomainNet, and consider multi-source domain adaptation, formalising for binary classification. They demonstrate multi-source training improves model accuracy, and publish baselines for state of the art methods.  %for image processing. Further, they propose Moment Distance between source and target domains, using feature extraction on each domain and sharing weights, as an input for a moment-matching classifier. Their formalism is defined only 

\subsection{Generalisability}
The language used in literature is not consistent. \citet{bareinboim2013general} highlights that generalisability goes under different "rubrics" such as external validity, meta-analysis, overgeneralisation, quasi-experiments and heterogeneity.

\citet{boulenger2005can} disambiguate terms in the context of healthcare economics (such as generalisability, external validity, and transferability), and created a self-reporting checklist to attempt to quantify transferability. They define generalisability as "the degree to which the results of a study hold true in other settings", and "the data, methods and results of a given study are transferable if (a) potential users can assess their applicability to their setting and (b) they are applicable to that setting". They advocate a user-centric view of transferability, considering specific usability aspects such as explicit currency conversion rates.

\citet{antonanzas2009transferability} create a transferability index at general, specific and global levels. Their "general index" is comprised of "critical factors", which utilise \citeauthor{boulenger2005can}'s factors, adding subjective dimensions.

\section{Transportability in NLP}

\subsection{Definitions}

To support a rigorous discussion, notational conventions are introduced. Extending the choices of \citet{pearl2011transportability}, we denote a domain $\mathcal{D}$ with population $\Pi$, governed by feature probability distribution $P$, which is data taken from a particular domain. We denote the \textit{source} with a $0$ subscript.

\begin{definition}
Generalisability: A system $\Psi$ has performance $p$ for solving task $T_0$ in domain $\mathcal{D}_0$. Generalisability is how the system $\Psi$ performs for solving task $T_i$ in domain $\mathcal{D}_j$, relative to the original task, without retraining. %$(\Psi,T_i,\mathcal{D_i}) \rightarrow (\Psi,T_j,\mathcal{D}_k)$
\end{definition}
Special cases, such as transportability or transference, have some $i,j=0$ in the definition above.
\begin{definition}
Transportability: A system $\Psi$ has performance $p$ for solving task $T_0$ in domain $\mathcal{D}_0$. Transportability is the performance of system $\Psi$ for solving task $T_0$ in a new domain $\mathcal{D}_i$, relative to the original task, without retraining. 
\end{definition}

Across multiple $\mathcal{D}_i$, we have relative performance $\tau_p(\mathcal{D}_0,\mathcal{D}_i)$, from which we can establish statistical measures for transportability performance and variation.

\begin{figure}
    \centering
    \includegraphics[scale=0.45]{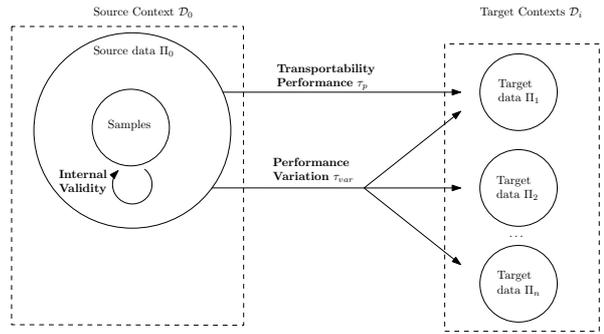}
    \caption{Schematic representation of the definitions}
    \label{fig:transportmap}
\end{figure}

Transfer learning is a specific way of achieving transportability (between populations or domains) or generalisability (including between tasks). \citet{singleton2014transfer} state that "transport encompasses transfer learning in attempting to use statistical evidence from a source on a target, but differs by incorporating causal assumptions derived from a combination of empirical source data and outside domain knowledge.". Note that this is different to \textit{generalisation} in the Machine Learning sense, which is akin to internal validity \citep{marsland2011machine}. Figure \ref{fig:transportmap} shows the definitions associated with transportability discussed in this paper.

%Recall Bareiboim and Pearl's definition:
%\begin{definition}
%Transportability: "A licence to transfer information learning in experimental studies to a different population, on which only observational studies can be conducted". 
%\end{definition}

%These definitions of transport and transfer are compatible with existing computer science terminology. For example, FTP is a protocol from transfer from one computer to another, with a transport layer (truck). 

%\iffalse
Table \ref{tab:terminology} summarises terminology, of how the target differs from source $(\Psi_0,T_0,\mathcal{D}_0(\Pi_0))$.
\begin{table}[ht]
    \centering
    \begin{tabular}{llllll}
    \hline
     \textbf{Term} & \textbf{$\Psi$} & \textbf{T} & \textbf{$\mathcal{D}$} & \textbf{$\Pi$} & \\
     \hline
     Cross-validation & 0 & 0 & 0 & i \\% & e.g. Table \ref{tab:evaluation_measures}\\ Internal validity
     New modeling & i & 0 & 0 & 0\\% & e.g. Table \ref{tab:evaluation_measures}\\
     %Internal Validity & i & i & j & i\\% & Section \ref{section:robustness}\\
     Transportability & 0 & 0 & i & i \\% & $\frac{\partial F_1}{\partial A}$, for a set of attributes $A$\\
     Transferability & 0 & 0,i & i & i \\
     Generalisability & 0 & 0,i & 0,i & 0,i \\% & Needed in sections\\ external validity
%     External Validity & A/B & B & B & A & \\
     \hline
    \end{tabular}
    \caption{Terminology through variation from a source. Table body is subscripts.}
    \label{tab:terminology}
\end{table}
%\fi

\textit{Chance}, \textit{bias} and \textit{confounding} are the three broad categories of "threat to validity". Broadly, chance and bias can be assessed by cross-validity, as it applies a model to the same task in the same domain on different data population. Confounding, error in interpretation of what is being measured, is more difficult to assess. %confounding is like "people who carry matches are more likely to get cancer - an error in interpretation of what is being measured.
Transportability is concerned with the transfer of learned information, with particular advances in the transport of causal knowledge.

Generalisability is the catch-all term for how externally valid a result or model is. Any combination of task, domain and data can be used. %Because of this, and of interest in the case of "Strong AI" it also includes the concept of generalising from an experiment to a theory.

\subsection{Transportability performance}

We define transportability performance $\tau_p$ as the gradient of the change in the performance metric's score from one domain to another. This measure does not take into account the underlying probability distributions, only the change in resulting performance measure.
\begin{equation}
\label{eqn:perf}
    \tau_p(\mathcal{D}_0,\mathcal{D}_i) = \frac{\textrm{p}(\Psi,T,\mathcal{D}_i)}{\textrm{p}(\Psi,T,\mathcal{D}_0)}
\end{equation}
\newline
The measure uses a ratio in order to allow comparison between different systems. To generalise this measure across different settings, we can take an average to give Equation \ref{eqn:perfgen}. Note that this is the average percentage change in performance, not an aggregated performance measure.

\begin{equation}
\label{eqn:perfgen}
    \tau_p(\mathcal{D}_0) = \frac{1}{n}\sum_{i=1}^n  \frac{\textrm{p}(\Psi,T,\mathcal{D}_i)}{\textrm{p}(\Psi,T,\mathcal{D}_0)}
\end{equation}

An analogous definition holds for different tasks over the same domain, $\tau_p(T)$.

\subsection{Performance variation}
\label{section:perfvariation}

Performance variation reflects how stable performance is across different contexts and can include, for example, to what extent the sampling method from the source data effects the performance metric of the algorithm. Part of this is data representativeness, the extent to which the source data representation also represents the target data.

More formally, performance variation $\tau_s(\Psi,T,\mathcal{D})$ is the change in performance of $(\Psi,T,\mathcal{D})$ across different contexts. This is useful in order to gain specific insight into external validity and generalisability. Indeed, we can assess the change in performance between source context $\mathcal{D}_0$ and target context $\mathcal{D}_i$. The source context has a privileged position, in that it is this space which the "learning" takes place, and the proposed metric for performance variation to multiple different domains is based on $\tau_p$ to reflect this.
Through repeated measurement in different contexts, we can go further. 
\begin{definition}
Performance Variation: For a model trained on domain $\mathcal{D}_0$ and applied on $n$ new domains $\mathcal{D}_i$, we define the performance variation as the coefficient of variation of performance across this set of domains so that:
\begin{equation}
\label{eqn:stab}
    \tau_{var}(\mathcal{D}_0) = \bigg(1+\frac{1}{4n}\bigg) \frac{\sqrt{\frac{\sum_{i=1}^n(\tau_p(\mathcal{D}_0,\mathcal{D}_i)-\tau_p(\mathcal{D}_0))^2}{n-1}}}{\tau_p(\mathcal{D}_0)}
\end{equation}

\end{definition}

The $1+\frac{1}{4n}$ term corrects for bias. In order to be meaningful, the target contexts must to have a good coverage of different domains. Enumerating these would be a task of ontological proportions, but can be pragmatically approximated by using the available Gold Standard datasets. 

We can also assess ability to generalise not just over different domains, but also different tasks, provided they can be meaningfully assessed by the same performance measure. We can consider $n$ different domain-task combinations, and with $\mean{\tau_{p}} = \sum_{i,j=0}^n \tau_{p}(\Psi,T_i,\mathcal{D}_j)/n$, this gives a more general form for Equation \ref{eqn:stab}, with $n$ large:
\newline
\begin{equation}
\label{eqn:stabgen}
    \tau_{var}= \frac{\sqrt{\frac{\sum_{i\geqslant 0,j \geqslant 0}(\tau_{p}(\Psi,T_i,\mathcal{D}_j)-\mean{\tau_{p}})^2}{n-1}}}{\mean{\tau_{p}}}
\end{equation}
In the case where different tasks cannot be assessed by the same measure, we are still able to compare different systems by looking at how the respective measures change.
 
\subsubsection{Performance variation properties}
For a purely random system, the transportability should be related to how similar the distributions of "answers" in the test dataset are. A random system should really be transportable by our measures. Similarly, we can consider trivial systems, such as identity and constant functions, which are necessarily entirely transportable. That is, for a system that is an identity function $\Psi=I$, $\tau_p=f(P)$, and $\tau_{var}(I,T,\mathcal{D}_i) = \tau_{var}(I,T,\mathcal{D}_j)=0$, $\forall i,j$. Note that we would not expect the same performance of these functions on different tasks.

A stable system will have $\tau_{var}(\Psi,T,\mathcal{D}_0) \approx \tau_{var}(\Psi,T,\mathcal{D}_i) \forall i$, reflecting that it is resilient to the domain on which it is trained.

\subsubsection{Factors influencing performance variation}

Through repeated measurement, we can quantify how $F_1$-score changes with respect to different measures $A$ (e.g. dataset complexity), $\frac{\partial F_1}{\partial A}$, with other properties held constant.

NLP system performance is dependent on $A$. This list may include gold standard feature distribution (in terms of representativeness of the semantic or linguistic phenomena), and task difficulty or sensitivity.

Users of NLP systems would benefit from being able to estimate the performance of an existing NLP system on a new domain, without performing the full implementation. Important for the performance of an NLP system, especially for few or zero shot learning, is having a common set of features (or phenomena) across domains. We proceed to propose three measures of increasing complexity, in order to attempt to understand how "similar" two domains are.

\paragraph{Lexical feature difference:} A measure grounded on lexical features (i.e. bag of words). The intuition behind this measure is for treating the set of lexical features as a representation. Linguistic space is observed as materialised tokens, which in turn are in some higher-dimensional semantic space, which enable interpretation. The measure considers the overlap of these linguistic spaces, and indeed the extent to which the linguistic space is covered by the data. Due to the simplicity of this measure, correlation between this and actual transportability performance is likely to be weaker than other measures but is simpler to calculate.
\begin{equation}
     \textrm{Lexical Feature Difference} = 1 - \frac{|\mathcal{D}_{i} \cap \mathcal{D}_0|}{|\mathcal{D}_i|} , i>0
\end{equation}
Where $|\mathcal{D}_i|$ is the number of features in the target domain $\mathcal{D}_i$, and $|\mathcal{D}_i\cap\mathcal{D}_0|$ is the number of features overlapping. This measure is then the proportion of unseen features in the new dataset. If all features of $\mathcal{D}_i$ are found in $\mathcal{D}_0$, then the feature difference is 0. If no features of $\mathcal{D}_i$ are found in $\mathcal{D}_0$, then the feature difference is 1. The feature overlap is task specific, and therefore appropriate to consider for transportability, but not generalisability. 

In the simplest case, the transported performance of a bag of words model should be precisely the lexical feature difference combined with distributions of the source and target domains. The feature set can range from binary lexical features to latent vector spaces. For different models, which target different aspects of semantic phenomena, different semantic and syntactic features will matter more. For this reason, considering a set of measures for domain complexity is warranted. In the context of this work, two measures are used over more complex feature spaces. 

\paragraph{Cosine distance:} Specifically, we use Doc2Vec \citep{le2014distributed} to embed the documents from each domain in a 300-dimensional feature-vector space, normalise, and calculate cosine distance to compare source and target domains.

\paragraph{Kullback–Leibler divergence:} Considering each domain as a distribution of features, we can use relative entropy to understand the difference between the source and target domains. Similar to cosine distance, we convert the corpus to a vector using Doc2Vec and normalize. We treat these values as discrete probability distributions to calculate the KL divergence.

% \begin{equation}
%     Z = \frac{X - min(X)}{max(X)-min(X)}
%     \label{eq:normalize}
% \end{equation}

%Alternative theoretical dataset complexity measure: How many rules would be required to perfectly describe the data? (Philip and VADER-inspired)

The usefulness of any of these domain similarity measures depends on the semantic phenomena and supporting corpora underlying the system, for example if the system requires a large training dataset, it may be more appropriate to use a measure which considers the underlying probability distributions in each feature. In this case, we can restrict to the case of the same task in order to keep the essential features reasonably consistent across domains. This makes this a measure of transportability (rather than generalisability). %Here, we propose using Kullback-Leibler divergence, or following Pearl(?). XXXX

There are additional dimensions of transportability potentially worthy of further investigation and quantification: (i) domain similarity (e.g. missing features), (ii) data efficiency (redundant/repeated features), (iii) data preparation (initial setup and formatting) and (iv) data manipulation required (data pipeline). %customisation. bespoke.

\section{Experiments}
\label{section:experiment}

\subsection{Setup}
The experiments aim to evaluate the consistency of the proposed transportability measures in the context of two standard tasks: named entity recognition and natural language inference. For reproducibility purposes the code and supporting data are available online\footnote{\url{https://github.com/ai-systems/transportability}}.

\begin{figure}
    \centering
    \includegraphics[width=\columnwidth]{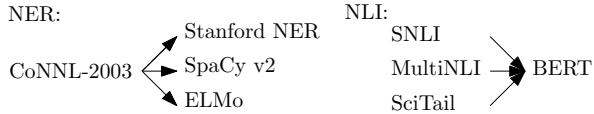}
    \caption{Overview of the experiments undertaken, indicating the models being applied to each dataset}
    \label{fig:experiment_setup_picture}
\end{figure}

We calculated the F1 score of multiple models on multiple datasets (Figure \ref{fig:experiment_setup_picture}). Note that in general the applicability of the proposed transportability measures are not limited to the use of F1 score, but this is simpler as the same measure applies for both tasks. All models and datasets are standard. For NER, the datasets were chosen as they have the consistent tags: Location, Person and Organisation. Stanford NER \citep{finkel2005incorporating} is a CRF classifier, SpaCy v2 is a CNN, ELMo \citep{ELMo} is a vector embedding model which outperforms GloVe and word2vec. Each of the three models used are trained on the CoNLL-2003 dataset~\cite{sang2003introduction}. We evaluated these models on CoNLL-2003, Wikipedia NER~\cite{ghaddar2017winer} (Wiki) and WNUT datasets~\cite{baldwin2015shared} for NER in twitter microposts.

For NLI, we chose to use standard datasets. SNLI \citep{snli:emnlp2015} is well established with a limited range of NLI statements, MultiNLI \citep{multinli} is multigenre with a more diverse range of texts, and SciTail \citep{khot2018scitail} is based on scientific exam questions. We applied BERT \citep{devlin2018bert}, a state of the art embedding model, to these datasets. 

\subsection{Results}

\begin{table}
\centering
\begin{tabular}{p{1cm}p{1cm}lp{1cm}p{1cm}} 
\toprule
\multicolumn{2}{c}{Dataset}&\multicolumn{2}{c}{Model} \\
\cmidrule{3-5}
 & & Stanford & SpaCy & ELMo  \\ 
 \midrule
\multirow{3}{*}{\parbox{1cm}{CoNLL-2003}} & Train                 & 98.69    & 99.32 & 99.97     \\ 
                       & Dev                   & 93.22    & 81.56 & 98.17     \\ 
                       & Test                  & 88.78    & 88.11 & 93.79     \\ 
\midrule
\multicolumn{2}{l}{Wiki}                     & 66.31    & 52.14 & 79.4      \\ 
\midrule
\multirow{3}{*}{WNUT}  & Train                 & 51.63       &  27.03    & 36.3      \\ 
                       & Dev                   & 53.59        & 32.23    & 48.8      \\ 
                       & Test                  & 47.11        & 26.28     & 58.1      \\
\bottomrule
\end{tabular}
\caption{NER F1 scores for different models trained on CoNLL dataset transported across different corpora}
\label{tab:NER}
\end{table}

% \begin{table}
% \centering
% \begin{tabular}{|p{1cm}|p{1cm}|l|p{1cm}|p{1cm}|} 
% \hline
% \multicolumn{2}{|l|}{\backslashbox[30mm]{Dataset}{Model}}                         & Stanford & SpaCy & Allen NLP  \\ 
% \hline
% \multirow{3}{*}{\parbox{1cm}{CoNLL-2003}} & Train                 & 98.69    & 99.32 & 99.97     \\ 
% \cline{2-5}
%                       & Dev                   & 93.22    & 81.56 & 98.17     \\ 
% \cline{2-5}
%                       & Test                  & 88.78    & 88.11 & 93.79     \\ 
% \hline
% \multicolumn{2}{|l|}{Wiki}                     & 66.31    & 52.14 & 79.4      \\ 
% \hline
% \multirow{3}{*}{WNUT}  & Train                 & 51.63       &  27.03    & 36.3      \\ 
% \cline{2-5}
%                       & Dev                   & 53.59        & 32.23    & 48.8      \\ 
% \cline{2-5}
%                       & Test                  & 47.11        & 26.28     & 58.1      \\
% \hline
% \end{tabular}
% \caption{NER F1 scores for different models trained on CoNLL dataset transported across different corpora}
% \label{tab:NER}
% \end{table}

\begin{table*}
%\ra{1.2}
\centering
\begin{tabular}{@{}p{4cm}cccccccc} 
\toprule
\multirow{3}{*}{\textbf{Source Dataset}} & \multicolumn{8}{c}{\textbf{Target Dataset}}\\
\cmidrule{2-9}
&\multicolumn{3}{l}{\makecell{SNLI~Dataset}} & \multicolumn{2}{l}{\makecell{MultiNLI Dataset}} & \multicolumn{3}{l}{\makecell{SciTail Dataset}}  \\ 
\cmidrule{2-9}
                  & Train & Dev & Test                & Train & Dev                           & Train & Dev & Test                    \\ 
\cmidrule{1-9}
SNLI (Train)              &  96.81     &   90.83  &  90.40                   &   72.51    &  72.29                             &   54.04    &  61.34   &     52.72                    \\ 
Multi NLI (Train)       &  77.13     &  79.05   &       79.31              &   97.78    & 83.50                          &  66.52     &  67.79   & 67.26                        \\ 
SciTail (Train)           &  42.68     &    44.36 &  44.20                   &    47.49   &                              44.49 & 99.88 &     94.78 &    93.08                          \\
\bottomrule
\end{tabular}
\caption{NLI accuracy scores for BERT model trained on one dataset transported to a different dataset}
\label{tab:NLI}
\end{table*}

\begin{table}
\centering
\begin{tabular}{llccp{1.2cm}} 
\toprule
\multicolumn{2}{l}{\textbf{Dataset}} &\textbf{Lexical}& \textbf{Cosine} & \textbf{KL Divergence}  \\ \midrule
\multirow{3}{*}{CoNLL} & Train & 0.000 & 0.000  & 0.000          \\ 

                      & Dev   & 0.121 &0.001  & 0.345          \\ 

                      & Test  & 0.197 &0.003  & 0.463          \\ 

\multicolumn{2}{l}{Wiki}     & 0.290 &0.007  & 0.701          \\ 

\multirow{3}{*}{WNUT}  & Train & 0.421 &0.134  & 2.129          \\ 

                      & Dev   & 0.511 &0.167  & 1.473          \\

                      & Test  & 0.481 &0.130  & 1.137          \\
\bottomrule
\end{tabular}
\caption{Domain similarity scores between the training corpus (CoNLL-2003) across other NER datasets}

\label{tab:ner_gradient_scores}
\end{table}

\begin{table*}
\centering
\resizebox{0.9\textwidth}{!}{\begin{tabular}{p{2cm}lllllllll} 
\toprule
\multirow{2}{*}{Dataset}  & \multirow{2}{*}{Measurement} & \multicolumn{3}{l}{\makecell{SNLI}} & \multicolumn{2}{l}{\makecell{MultiNLI}} & \multicolumn{3}{l}{\makecell{SciTail}}  \\ 
\cmidrule(lr){3-5}  \cmidrule(lr){6-7}
\cmidrule(lr){8-10}
                          &                              & Train & Dev   & Test      & Train & Dev                   & Train & Dev   & Test          \\ 
\midrule
\multirow{2}{2cm}{SNLI (Train)}  & Lexical & 0.000 & 0.003 & 0.003 & 0.086 & 0.088 & 0.136 & 0.115 & 0.119  \\
    & Cosine                        & 0.000 & 0.002 & 0.002     & 0.008 & 0.007                 & 0.233 & 0.242 & 0.242         \\ 
                          & KL Divergence         & 0.000 & 3.277 & 4.283   & 6.489   & 8.982                 & 16.02 & 17.50 & 18.20       \\
\midrule
\multirow{2}{2cm}{MultiNLI (Train)} & Lexical  & 0.008 & 0.008 & 0.008 & 0.000 & 0.008 & 0.063 & 0.063 & 0.047  \\

& Cosine       & 0.008 & 0.018 & 0.016     & 0.000 & 0.002                 & 0.298 & 0.307 & 0.306         \\ 
                          & KL Divergence         & 11.07 & 7.613 & 6.333   & 0.000 & 3.342                & 33.10 & 35.27 & 34.69         \\
\midrule
\multirow{2}{2cm}{SciTail (Train)} & Lexical & 0.282 & 0.282 & 0.282 & 0.277 & 0.278 & 0.000 & 0.028 & 0.025  \\
& Cosine       & 0.233 & 0.230 & 0.231     & 0.262 & 0.298                 & 0.000 & 0.001 & 0.002        \\
                          & KL Divergence         & 11.17 & 7.04 & 7.492     & 5.220 & 6.682                 & 0.000  & 1.097 & 1.424          \\
\bottomrule
\end{tabular}}
\caption{Domain similarity scores between the source training corpus and target corpora}
\label{tab:nli_gradient_scores}
\end{table*}

\begin{figure*}[ht]
    \centering
    \captionsetup{justification=centering,margin=1cm}

    \begin{subfigure}[b]{0.4\linewidth}
    
\begin{tikzpicture}[scale=.65]
    \centering
\begin{axis}[
    xlabel={Cosine Distance $\times 10^{-2}$},
    ylabel={F1 Score},
    xmin=0, xmax=17,
    ymin=0, ymax=100,
   % legend style={at={(0.5,-0.1)},anchor=north}    % xtick={97,96,93,79,54,58},
    % ytick={0,1,3,7,167,134,130},
    legend pos=south west,
    ymajorgrids=true,
    grid style=dashed,
]
 
\addplot[
    color=blue,
    mark=triangle*,
    ]
    coordinates {
    (0, 97)(0.1,96)(0.3,93)(0.7,79)(13, 58)(13.4,67)(16.7, 54)
    };

\addplot[
    color=red,
    mark=triangle*,
    ]
    coordinates {
    (0, 94)(0.1,93)(0.3,89)(0.7,66)(13, 53.69)(13.4,51.63)(16.7, 47.11)
    };

\addplot[
    color=brown,
    mark=triangle*,
    ]
    coordinates {
    (0, 99.32)(0.1,88.78)(0.3,81.5)(7,52.14)(13, 32.23)(13.4,27.3)(16.7, 26.28)
    };
     \legend{ELMo NER, Stanford NER, SpaCy NER}
\end{axis}
\end{tikzpicture}
\caption{NER F1 scores Vs Doc2Vec cosine distance from training (CoNLL) corpus}
\label{fig:ner_cos_divergence}
\end{subfigure}
\hspace{15mm}
\begin{subfigure}[b]{0.4\linewidth}
\begin{tikzpicture}[scale=.65]
\begin{axis}[
    xlabel={KL Divergence},
    ylabel={F1 Score},
    xmin=0, xmax=2.5,
    ymin=0, ymax=100,
  %  legend style={at={(0.5,-0.1)},anchor=north}    % xtick={97,96,93,79,54,58},
    % ytick={0,1,3,7,167,134,130},
    legend pos=south west,
    ymajorgrids=true,
    grid style=dashed,
]
 
\addplot[
    color=blue,
    mark=triangle*,
    ]
    coordinates {
    (0, 99)(0.345,96)(0.463,93)(0.701,79)(1.137, 58.1)(1.473,48.8)(2.129, 36.3)
    };
    \addplot[
    color=red,
    mark=triangle*,
    ]
    coordinates {
    (0, 98.69)(0.345,93.22)(0.463,88.78)(0.701,66.31)(1.137, 53.59)(1.473,51.63)(2.129, 47.11)
    };
    
    \addplot[
    color=brown,
    mark=triangle*,
    ]
    coordinates {
    (0, 99.31)(0.345,81.46)(0.463,88.11)(0.701,52.64)(1.137, 32.23)(1.473,27.03)(2.129, 26.28)
    };
    
     \legend{ELMo NER, Stanford NER, SpaCy NER}
\end{axis}
\end{tikzpicture}
\caption{NER F1 scores Vs KL Divergence from training (CoNLL) corpus}
\label{fig:ner_kl_divergence}
\end{subfigure}

% \begin{subfigure}[b]{0.4\linewidth}
% \begin{tikzpicture}[scale=.7]
% \begin{axis}[
%     xlabel={Lexical Similarity $\times 10^{2}$},
%     ylabel={F1 Score},
%     xmin=0, xmax=240,
%     ymin=0, ymax=100,
%     legend style={at={(0.5,-0.1)},anchor=north}    % xtick={97,96,93,79,54,58},
%     % ytick={0,1,3,7,167,134,130},
%     legend pos=north west,
%     ymajorgrids=true,
%     grid style=dashed,
% ]
 
% \addplot[
%     color=blue,
%     mark=triangle*,
%     ]
%     coordinates {
%     (0, 99)(12.1,96)(19.7,93)(29,79)(42.1, 36.3)(48.1, 58.1)(51.1,48.8)
%     };
%     \addplot[
%     mark=triangle*,
%     ]
%     coordinates {
%     (0, 98.69)(12.1,93.22)(19.7,88.78)(29,66.31)(48.1, 53.59)(36.3,51.63)(51.1, 47.11)
%     };
    
%     \addplot[
%     color=brown,
%     mark=triangle*,
%     ]
%     coordinates {
%     (0, 99.31)(34.5,81.46)(46.3,88.11)(70.1,52.64)(113.7, 32.23)(147.3,27.03)(212.9, 26.28)
%     };
    
%      \legend{Allen NER, Stanford NER, SpaCy NER}
% \end{axis}
% \end{tikzpicture}
% \caption{NER F1 scores Vs Lexical similarity with training (CoNLL) dataset}
% \label{fig:ner_lexical_divergence}
% \end{subfigure}
\caption{NER F1 score plotted against different measures of corpus similarity}
\end{figure*}
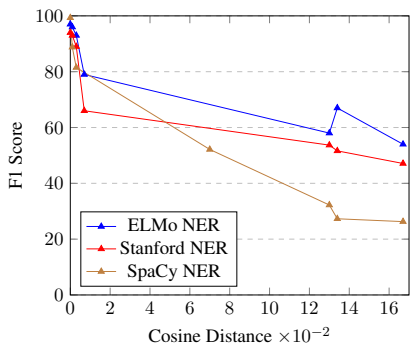
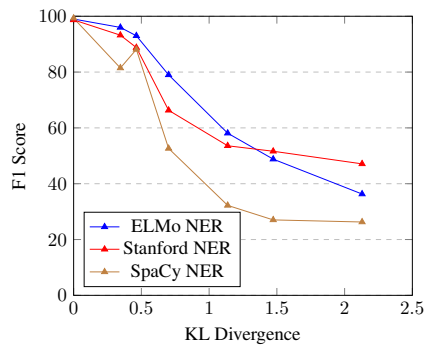

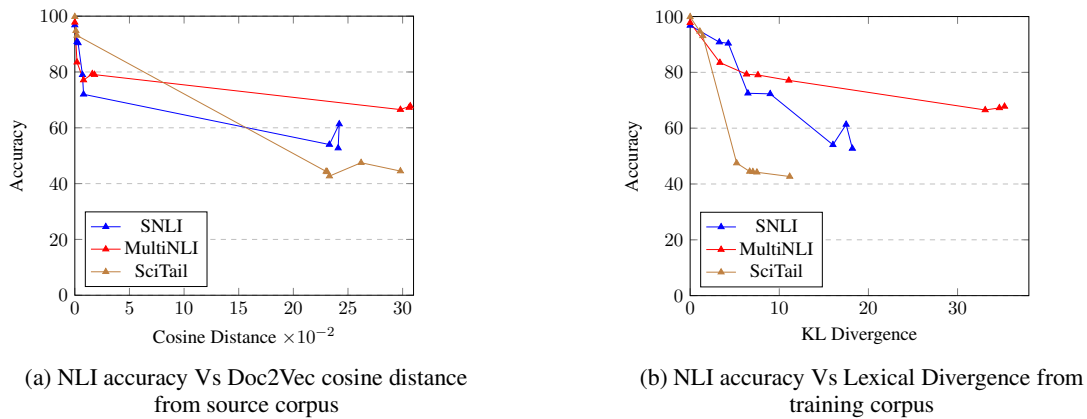
\begin{figure*}[ht]
    \centering
    \captionsetup{justification=centering,margin=1cm}

    \begin{subfigure}[b]{0.4\linewidth}
    
\begin{tikzpicture}[scale=.65]
\begin{axis}[
    xlabel={Cosine Distance $\times 10^{-2}$},
    ylabel={Accuracy},
    xmin=0, xmax=31,
    ymin=0, ymax=100,
  %  legend style={at={(0.5,-0.1)},anchor=north,nodes={scale=0.5, transform shape}}    % xtick={97,96,93,79,54,58},
    % ytick={0,1,3,7,167,134,130},
    legend pos=south west,
    ymajorgrids=true,
    grid style=dashed,
]
 
\addplot[
    color=blue,
    mark=triangle*,
    ]
    coordinates {
    (0, 96.81)(0.2,90.8)(0.3,90.4)(0.7, 79)(0.8,72)(23.3,54)(24.2, 61.34)(24.1, 52.72)
    };

\addplot[
    color=red,
    mark=triangle*,
    ]
    coordinates {
    (0,97.78)(0.2,83.5)(0.8, 77.1)(1.6,79.32)(1.8,79.05)(29.8,66.52)(30.7, 67.79)(30.6,67.26)
    };

\addplot[
    color=brown,
    mark=triangle*,
    ]
    coordinates {
    (0,99.8)(0.1,94.78)(0.2,93.08)(23,44.36)(23.1,44.2)(23.3,42.68)(26.2,47.49)(29.8,44.49)
    };
     \legend{SNLI, MultiNLI, SciTail}
\end{axis}
\end{tikzpicture}
\caption{NLI accuracy Vs Doc2Vec cosine distance from source corpus}
\label{fig:nli_cos_divergence}
\end{subfigure} 
\hspace{15mm}
\begin{subfigure}[b]{0.4\linewidth}
\begin{tikzpicture}[scale=.65]
\begin{axis}[
    xlabel={KL Divergence},
    ylabel={Accuracy},
    xmin=0, xmax=38,
    ymin=0, ymax=100,
    % legend style={at={(0.5,-0.1)},anchor=south}    % xtick={97,96,93,79,54,58},
    % ytick={0,1,3,7,167,134,130},
    legend pos=south west,
    ymajorgrids=true,
    grid style=dashed,
]
 
\addplot[
    color=blue,
    mark=triangle*,
    ]
    coordinates {
    (0, 96.81)(3.27,90.83)(4.28,90.4)(6.48,72.51)(8.98, 72.29)(16.02,54.04)(17.50, 61.34)(18.20, 52.72)
    };
    
    \addplot[
    color=red,
    mark=triangle*,
    ]
    coordinates {
    (0,97.78)(3.34,83.5)(6.33,79.31)(7.61,79.05)(11.07,77.13)(33.10,66.52)(34.69,67.26)(35.27,67.79)
    };
    
    \addplot[
    color=brown,
    mark=triangle*,
    ]
    coordinates {
    (0,99.88)(1.09,94.7)(1.42,93.08)(5.22,47.49)(6.68,44.49)(7.04,44.36)(7.49,44.2)(11.17,42.68)
    };
    
     \legend{SNLI, MultiNLI, SciTail}
\end{axis}
\end{tikzpicture}

% \begin{tikzpicture}[scale=.65]
% \begin{axis}[
%     xlabel={Lexical Divergence $\times 10^{3}$},
%     ylabel={Accuracy},
%     xmin=0, xmax=500,
%     ymin=0, ymax=100,
%     % legend style={at={(0.5,-0.1)},anchor=south}    % xtick={97,96,93,79,54,58},
%     % ytick={0,1,3,7,167,134,130},
%     legend pos=south east,
%     ymajorgrids=true,
%     grid style=dashed,
% ]
 
% \addplot[
%     color=blue,
%     mark=triangle*,
%     ]
%     coordinates {
%     (0, 96.81)(3,90.83)(3,90.4)(86,72.51)(88, 72.29)(115, 61.34)(119, 52.72)(136,54.04)
%     };
    
%     \addplot[
%     color=red,
%     mark=triangle*,
%     ]
%     coordinates {
%     (0,97.78)(8,83.5)(8,79.31)(8,79.05)(8,77.13)(47,67.26)(63,66.52)(63,67.79)
%     };
    
%     \addplot[
%     color=brown,
%     mark=triangle*,
%     ]
%     coordinates {
%     (0,99.88)(25,93.08)(28,94.7)(277,47.49)(278,44.49)(282,44.36)(282,44.2)(282,42.68)
%     };
    
%      \legend{SNLI, MultiNLI, SciTail}
% \end{axis}
% \end{tikzpicture}

\caption{NLI accuracy Vs Lexical Divergence from training corpus}
\label{fig:nli_kl_divergence}
\end{subfigure}

\caption{NLI accuracy score plotted against different measures of corpus similarity}
\label{fig:nli_divergence}
\end{figure*}

\paragraph{NER:}Table \ref{tab:NER} shows results for the NER task, trained on CoNLL. Unsurprisingly, all models performed better when the target was in the CoNLL domain. The reduced performance on Wiki was more extreme than expected, particularly for ELMo, which was expected to be resilient to domain change (i.e. transportable). Table~\ref{tab:NER_Transport} and Table~\ref{tab:ner_gradient_scores} illustrate the transportability and domain similarity scores for different NER models respectively.

\paragraph{NLI:} Table \ref{tab:NLI} shows results for the NLI task, using BERT. We find that, despite the vast training data, BERT's performance is substantially higher when it has been trained on data from that domain. BERT trained on SciTail performs poorly when transported to SNLI or MultiNLI.  Table~\ref{tab:NLI_Transport} and Table~\ref{tab:nli_gradient_scores} illustrates the transportability and domain similarity scores for different NLI corpora.

\subsection{Analysis}

Every model had $\tau_p\ll 1$, meaning they performed worse on the new domain. This is as expected, though this would not be true in general. 

\paragraph{NER:} Examining the F1 scores (88.11 vs. 88.78) of SpaCy and Stanford they appear almost comparable. However, the latter transports much more effectively, with $\tau_p$ score difference (0.671 Vs 0.524 when transporting to Wiki) (refer Table~\ref{tab:NER_Transport}).

ELMo is one of the state of the art approaches for NER, as evidenced by the high F1 scores for the source corpus. However, Stanford NER transports equally well, and when transported outperforms ELMo for twitter domain. While the absolute F1 score difference between them is 5, the $\tau_p$ scores are almost identical, with a difference of 0.003. In terms of transportability, it is notable that an approach that employs CRF tagger with linguistic features outperforms significantly the CNN-based SpaCy approach and stands in comparison to a computationally expensive model like ELMo.

Stanford NER also has the lowest $\tau_{var}$. This indicates this to be the most robust model out of the three. This conclusion was facilitated by the $\tau_p$ and $\tau_{var}$ measures.

NER for English is assumed to be an accomplished task as supported by the traditional F1 scores. By using $\tau_p$ we argue that there is a need for more robust models, with better transportability performance.

Figure~\ref{fig:ner_cos_divergence} and Figure~\ref{fig:ner_kl_divergence} illustrates the decrease in F1 scores as cosine distance and KL divergence increase. A simple 3 parameter non-linear regression model on KL Divergence and Cosine distance is able to predict the F1 score with an mean error of 3.33 and 2.66 respectively. Considering the lexical difference has similar results (Table \ref{tab:ner_gradient_scores}). This implies that by using these measures we may be able to anticipate the accuracy of a model in a new domain based on easy to compute domain similarity, which is straightforward to compute.

% In NER, data with a higher cosine distance from the training data (CoNLL-2003) were found to have a lower F1 score. We would not expect, and did not find, a strong relationship, but it appears that cosine distance $>0.1$ many have a much lower performance. 

\paragraph{NLI:} Applying BERT to different domains was not as resilient to domain transport as we expected. The average $\tau_p$ is 0.612 over transported domains, despite these being standard corpora from the domains. We found MultiNLI(Train) to be more transportable than the others, since its performance in new domains is not much worse than new data from the same domain. This is as expected, since MultiNLI has been built to have good domain coverage. Specifically, MultiNLI has $\tau_p=0.744$ and $\tau_{var}=8.582$, whilst SNLI has $\tau_p=0.646$ and $\tau_{var}=15.22$ and SciTail has $\tau_p=0.446$ and $\tau_{var}=3.921$. SciTail transports poorly, and does so reliably! SNLI transports in between, but variably, being quite "hit or miss" with different samples of SciTail. 
These results suggest a threshold for $\tau_p$ of perhaps 0.8 as being "appropriate" for transportability performance. A threshold for $\tau_{var}$ is more difficult to establish and would benefit from further investigation. Clearly, these measures depend on the domains chosen. 

As with NER, we found lexical difference indicative of transported performance, and that for NLI, accuracy scores decrease with increasing lexical difference, cosine distance and KL divergence (Tables~\ref{tab:NLI} and ~\ref{tab:nli_gradient_scores}, and Figures ~\ref{fig:nli_cos_divergence} and ~\ref{fig:nli_kl_divergence}). A simple 3 parameter non-linear regression model on KL Divergence and Cosine distance is able to predict the accuracy score with an mean error of 3.98 and 1.95 respectively.

\subsection{Discussion}
\paragraph{$\tau_p$ and $\tau_{var}$ as complementary to traditional measures.} We are not breaking new ground in terms of evaluation methodology, but the experiments demonstrate that traditional F1 and accuracy measures do not capture a complete picture. Transportability measure are not only simple enough to calculate and convey but also evaluates a model with regards to generalisability and robustness.

% \paragraph{Domain similarity measures to open the black box.} Many modern models are black box, rendering it hard to understand `what if'. Although the domain similarity measures we have introduced are in trivial in nature, it shows a definite correlation, which could lead the way to peek into the black boxes and perhaps understand more about causal transportability.

\paragraph{Low cost ways of anticipating performance for a new task or domain.} Most of the state of the art models are computationally expensive. With the transportability and domain similarity measures we are able to predict performance in a new domain with reasonable accuracy. These similarity measures are relatively simpler to run.

% \paragraph{Non-neural architectures} Whilst not explored here, this method could also apply to models using structurally annotated data such as treebanks. %is this right?

% Discussion points:
%F1 isn't enough. This paper aims to start a conversation about this. 
%In addition to transportability $\tau_p$ is also useful for measuring internal validity
% what data will my model work on? We've shown that data similarity is useful for understanding transportability

\begin{table}
\centering
\resizebox{\columnwidth}{!}{\begin{tabular}{p{2.5cm}lll}
\toprule
 & Stanford & SpaCy & ELMo  \\
\midrule
$\tau_{p}($\textit{wiki}$)$ &    0.671&  0.524     &  0.794        \\
$\tau_{p}($\textit{wnut}$)$ &    0.514      & 0.287      & 0.477       \\
$\tau_{p}($\textit{wnut \& wiki}$)$ &     0.553     &   0.346    &   0.556      \\
\midrule
$\tau_{var}$& 15.051      & 35.171&  32.666       \\
\bottomrule
\end{tabular}}
\caption{Transportability measures for NER models}
\label{tab:NER_Transport}
\end{table}

\begin{table}
\centering
\begin{tabular}{lrrr} 
\toprule
&SNLI & MultiNLI & SciTail\\
\midrule
$\tau_{p}$                     & 0.646                                                                 & 0.744                                                                      & 0.446                                                                       \\ 
$\tau_{var}$                  & 15.22                                                                 & 8.582                                                                      & 3.921                                                                       \\
\bottomrule
\end{tabular}

\caption{Transportability measures for NLI corpora}
\label{tab:NLI_Transport}
\end{table}

\section{Conclusion}
We have presented a model of transportability for NLP tasks, together with metrics to allow for the quantification in the change in performance. We have shown that the proposed transportability measure allows for direct comparison of NLP systems' performance in new contexts. Further, we demonstrated domain similiarity as a measure to model corpus and domain complexity, and predict NLP system performance in unseen domains. This paper lays the foundations for further work in more complex transportability measures and estimation of NLP system performance in new contexts. 

\bibliographystyle{acl_natbib}
\bibliography{iwcs2021}

%\appendix

\end{document}